# A Factor Graph Model of Trust for a Collaborative Multi-Agent System

Behzad Akbari, Member, IEEE, Mingfeng Yuan, Student Member IEEE, Hao Wang, Haibin Zhu, Senior Members, IEEE, Jinjun Shan, Senior Members, IEEE

*Abstract* — In the field of Multi-Agent Systems (MAS), known for their openness, dynamism, and cooperative nature, the ability to trust the resources and services of other agents is crucial. Trust, in this setting, is the reliance and confidence an agent has in the information, behaviors, intentions, truthfulness, and capabilities of others within the system. Our paper introduces a new graphical approach that utilizes factor graphs to represent the interdependent behaviors and trustworthiness among agents. This includes modeling the behavior of robots as a trajectory of actions using a Gaussian process factor graph, which accounts for smoothness, obstacle avoidance, and trust-related factors. Our method for evaluating trust is decentralized and considers key interdependent sub-factors such as proximity safety, consistency, and cooperation. The overall system comprises a network of factor graphs that interact through trust-related factors and employs a Bayesian inference method to dynamically assess trust-based decisions with informed consent. The effectiveness of this method is validated via simulations and empirical tests with autonomous robots navigating unsignalized intersections.

## I. Introduction

In various applications of Multi-Agent Systems (MAS) like self-driving cars, e-commerce, and virtual worlds, the behavior of agents relies on observations or services provided by sensors or other agents. Behavior of the robot is the term used to describe a set of actions undertaken by agents with the objective of achieving mission goals within a specific context in some work named Process Role [1]. Agents are assumed to be rational and driven to maximize the performance of their team. However, in cooperative scenarios, there is a risk of agents receiving inaccurate information, which can disrupt the smooth operation of the system.

Trust, in this context, can be defined as the reliance and confidence that one agent places in the behaviors, intentions, truthfulness, and capabilities of other agents within the system. Trust encompasses multiple dimensions that characterize the relationship between agents, such as consistency, performance, communication, reliability, adaptability, cooperation, transparency, and reputation.

The concept of trust in MAS is complex for various reasons. Firstly, trust is context-dependent, meaning that it can vary based on the specific situation or environment in which agents operate. Secondly, trust is subjective and can differ among agents based on their roles or individual perspectives. Moreover, trust involves multiple factors and is characterized by its inherent complexity. Uncertainty and limited information further add to the challenges of establishing trust. Trust is also dynamic in nature, as it can evolve over time due to changing circumstances and interactions. Additionally, trust is influenced by the interactions and dependencies among agents, where trust in one agent may depend on the trustworthiness of others.

Human beings establish trust by observing past interactions, making new observations, or relying on testimonies. In a multi-agent system, agents can evaluate each other's actions by considering observed behavior and the trust-related factors. Trust can be modeled by assessing the probability of receiving false information or being deceived by an interaction partner. The evaluation of an agent's trustworthiness can be facilitated by taking into account historical interaction outcomes, new observations, and testimonies provided by other agents as direct evidence [2, 3].

Trust was first introduced as a quantifiable entity in computer science in a study [4]. Subsequently, various computational models on different aspects of trust management have been proposed. In earlier research, trust management was categorized into two sections, namely, trust evaluation and trust-aware decision-making or trust establishment [5, 6]. In most open and dynamic multi-robot systems, trust evaluation needs to be combined with trustworthy decision-making. For an accurate evaluation of trust, it is necessary to use context-based, real-time, robust, stochastic, and flexible models. Trust evaluation can be conducted based on trust models that rely on past behaviors and third-party testimonies from other agents in the environment [7, 8, 9, 10]. One approach for evaluating trust is to establish a central unit to collect and analyze observations using a computation facility like a cloud. However, this solution does not address the root cause of mistrust and creates new issues of inefficiency, bottlenecks, and information asymmetry, among others. Hence, it is crucial to have a centralized engine to overcome these problems.

In our paper, we present a graphical model that investigates the interplay of trust-related behaviors among agents. We utilize Bayesian inference to evaluate and synthesize multiple factors: those necessary for path planning and those critical for establishing trust. These include prior factors,

B. Akbari, and H. Zhu are with Collaborative Systems Laboratory (CoSys Lab), Department of Computer Science and Mathematics, Nipissing University, North Bay, Ontario, Canada (Email: {behzada, haibinz }@nipissingu.ca)

Mingfeng Yuan, Hao Wang, and Jinjun Shan are with the Department of Earth and Space Science, Lassonde School of Engineering, York University, ON M3J1P3, Canada (Email: {mfyuan, hwang12, jjshan}@yorku.ca)

smoothness, obstacle avoidance, and interdependent factors such as proximity safety, cooperation, transparency, and reputation. This approach enables us to concurrently estimate trust and take trustworthy action with consent. The assessment of an agent's trustworthiness depends on its behavior, underscoring the significance of trust development. Thanks to optimization through factor graphs, our proposed model efficiently optimizes collaborative and reliable actions. We represent the behavior of all robots as dependent stochastic processes in a multi-dimensional factor graph, which encompasses both independent and interdependent factors. This methodology facilitates consensus through Bayesian inference. Trust is established by generating dependable behaviors based on local observations and shared testimonies, with adjustments made in response to the evaluated trust factor.

The main contributions of this paper are as follows:
1. This paper introduces a new, versatile graphical model based on factor graphs, which is designed to simultaneously represent the trust and trustworthy behavior of agents within a single optimization problem.
2. We have implemented a general Bayesian inference approach using the factor graph model of agents and interactions to estimate the trust factor and establish trust. This method provides a deeper understanding of agent interactions, optimized behaviors, and consensus within a single framework.
3. The practical implementation of our proposed graphical model in a trust management system is demonstrated through its application to an unsignalized intersection problem.

**Table I:** Abbreviations adopted in this manuscript.

| Abbreviation | Description |
| --- | --- |
| MRS | Multi-Robot Systems |
| AI | Artificial Intelligence |
| GP | Gaussian Process [1] |
| GPMP | Gaussian Process Motion Planning [2] [3] |
| BRS | Beta Reputation System [13, 14, 15] |
| MAP | Maximum A Posteriori |
| LTV-SDE | Linear time-varying stochastic differential equation |
| BFT | Byzantine Fault Tolerance |
| HMM | Hidden Markov Model |
| DLT | Distributed Ledger Technology |
| SDF | Signed Distance Field |
| UAV | Unmanned Aerial Vehicle |
| EPD | Exponential Power Distribution |

## II. RELATED WORKS

*A. Previous Methods of Trust Modeling*

The Beta Reputation System (BRS) [4, 5, 6] is an early model that determines trustworthiness using direct evidence. BRS calculates trust by averaging the positive and negative feedback received by the trustee agent. The model can also incorporate uncertainty and forgetting factors to gradually discard evidence [7]. In [8], a multi-dimensional trust model is proposed, considering various factors such as success, budget, deadline, and quality likelihood to evaluate an agent's trustworthiness. However, when there is low observability and direct evidence is insufficient, indirect evidence and third-party testimonies can be used to estimate trust. But, biased testimonies can negatively impact trust-aware decisions. To weigh direct and indirect factors, a static or dynamic weight γ can be used. In [26], an approach based on the Q-learning technique was proposed to select an appropriate γ value from a predetermined static set of values. Additionally, in [9], fuzzy cognitive maps (FCMs) were suggested that use internal and external factors. Truster agents are able to calculate trust values by analyzing the relationships between various independent and interdependent factors to determine their preferred actions. However, inconsistencies in the sources of belief and the selection of values for these causal links can significantly impact the accuracy and reliability of the trust model. The first Bayesian inference model for trust management was introduced in [10]. They used confidence level and mapped the trusted network into a Bayesian Network in a very basic way. The confidence lower and upper values are used as heuristics to calculate the most accurate estimations of the trustworthiness of the trustee agents. Several other Bayesian methods based on Hidden Markov Model (HMM) were introduced in [11] [12] for trust management. In those works, the target agents' behavior is predicted according to the HMM trust estimation module following the Q-learning greedy policy. A complete reinforcement learning-based approach for trust management was proposed in [13]. The gain derived by a truster agent from choosing each trustee agent for interaction consists of the Q-value from Q-learning. A truster agent selects an action to maximize its gain at each step.

Most previous methods attempt to estimate or track an agent's trust value individually, based on prior behaviors and new observations. However, these methods, which assume mostly agents' independency, often do not yield satisfactory results in cooperative scenarios due to the actual dependencies among agents [14]. A more realistic approach involves considering both independent and interdependent factors for decision-making. In our paper, we propose a generic stochastic graph-based model that defines trustworthy behavior based on various singular and interdependent factors. We simplify the relationship of these stochastic variables for each agent using a multi_dimentional factor graph, Fig. 2. Bayesian inference is employed for decision-making. Additionally, our optimization method enforces consensus by considering the behavior of all agents within a general factor graph optimization framework.

*B. Variable-Based Modeling and GP*

In Artificial Intelligence (AI), variable-based modeling is an advanced methodology that employs probabilistic methods to solve various complex problems, especially those involving uncertainty. In this approach, probabilistic variables, which are unknown and need estimation, play a central role. AI's variable-based models predominantly use graphical models, namely Bayesian networks (Bayesnets) and factor graphs. Factor graphs are often the preferable option when dealing with numerous constraints and satisfactions. As a type of probabilistic graphical model, factor graphs are particularly effective for issues involving stochastic variables. They represent the conditional dependencies between variables within a network, facilitating efficient computation of joint probabilities and inferences. In these models, stochastic variables are interconnected. Bayesian methods are employed to update beliefs about these variables based on observed data, enabling dynamic and probabilistic reasoning about the system's state [15].

Gaussian Process (GP) is a stochastic process that models a distribution over functions by representing a continuous-time function as a small number of vector value states [1]. Inspired by variable-based models in AI, we can define a finite set of stochastic variables to simulate agent behavior using factor graph representation. A simple trajectory model for a Unmanned Aerial Vehicle (UAV) robot is illustrated in Fig. 1.The general method for solving such a model involves finding the posterior function of the target variables using Bayes' formula. Thanks to factor graphs combined with the sum-product algorithm and sparsity, we can efficiently and linearly compute posterior distributions. In [2], GP was combined with a factor graph for motion planning, creating Gaussian Process Motion Planning (GPMP). In our paper, we model the behaviour of each agent using a GP and its factor graph reprsentation. The GP factor graph allows for the use of Bayesian inference, to optimize context-based trustworthy actions or process roles.

In our approach, we take into account both regular factors related to path planning and inter-agent factors that influence robot behavior. Regular factors include aspects such as obstacle avoidance and smoothness. Additionally, we consider interdependent factors related to relationship dynamics and trust, ensuring a comprehensive understanding of the robots' interactions. For inference and variable determination in our model, we utilize the well-known Maximum A Posteriori (MAP) estimator. To efficiently handle real-time computation, we leverage sparsity. This approach allows us to transform the MAP problem into a linear equation, enabling timely and effective solutions.

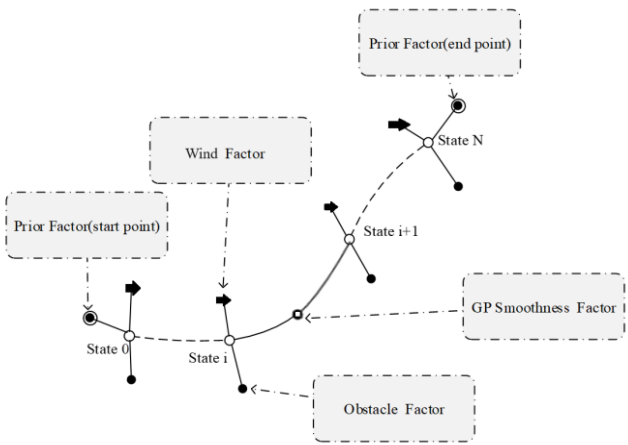

**Fig. 1.** A factor graph of a trajectory for a UAV robot path planning. Note that the support variables (states) are marked as white circles, and there are four main types of factors, namely prior factors on the start and goal states, smoothness (GP) factors, obstacle factors and wind factor for UAV.

*C. Consensus-Based Decision making*

Achieving consensus among agents in collaborative multi-agent systems poses a fundamental challenge. To attain an agreement, agents must exchange information to generate control laws, ensuring a consistent view of their states. However, conventional methods often rely on a central authority for the consensus algorithm, leading to reduced network reliability. The consensus mechanism requires access to other agent information and the ability to recover from potential contributing node failures, ensuring a certain level of fault tolerance [16] [17] [18]. Autonomous multi-robot systems face risks associated with inadequate direct observations or biased testimonies, which can impact decision-making. To mitigate these challenges, Distributed Ledger Technology (DLT), like Blockchain, can be utilized to offer transparency, traceability, and security. DLT utilizes a temper-proof ledger distributed among all participants, safeguarding data integrity. Notably, consensus algorithms remain a crucial aspect of distributed ledgers, such as the widely known Byzantine Fault Tolerance (BFT) algorithm involving pre-prepare, prepare, and commit steps.

Another well-known consensus method used in decision-making is the Bayesian consensus algorithm. Bayesian consensus refers to a decision-making process that incorporates Bayesian reasoning and principles. Bayesian reasoning is a mathematical and philosophical approach that involves updating beliefs based on new evidence. In the context of consensus, Bayesian reasoning can be applied to situations where multiple individuals or agents need to make a collective decision or reach an agreement. The process involves iteratively updating individual beliefs or probabilities based on new information, which could be in the form of data, observations, or opinions from other participants. Bayesian consensus starts with an initial belief, sharing information, and progressively updates beliefs over time, similar to [19] and [16]. By employing Bayesian consensus, control laws can be generated, ensuring convergence to a consistent view of agents' beliefs. This convergence occurs because participants are incorporating both their own information and the information from others. In our paper, we employed a typical distributed network, focusing on exceptions related to optimizing trustworthy behavior. The Bayesian consensus was used to amalgamate agents' streams of actions, taking into account inter-dependent factors.

III. PROBLEM FORMULATION

We utilized Gaussian Process (GP) factor graphs, as illustrated in Fig. 1, to model each agent's process role. The connection between each GP factor graph is managed using Prior, Unary and Trust-related factors, as shown in Fig. 2. All factors have been employed to identify trustworthy and optimized behavior. Our proposed model facilitates both the optimization of behavior and the establishment of trust. For ordinary factors, we consider aspects like GP prior, and obstacle avoidance factors. For trust-related factors, we focus on interdependent elements such as proximity safety, cooperation, and transparency. Our inference approach combines the trust factor with these other factors to determine dependent trajectories that maximize trust.

*A. Mathematical Model*

To optimize the trustworthy behavior among agents, we have approached it as a generic optimization problem. We calculate trust using sub-factors. The assessment of trust between two agents, truster i (the party placing trust) and trustee j (the party being trusted), involves examining how their behaviors are interrelated. Our methodology divides elements for evaluating the trust factor into three main categories, proximity safety, cooperation, and transparency.

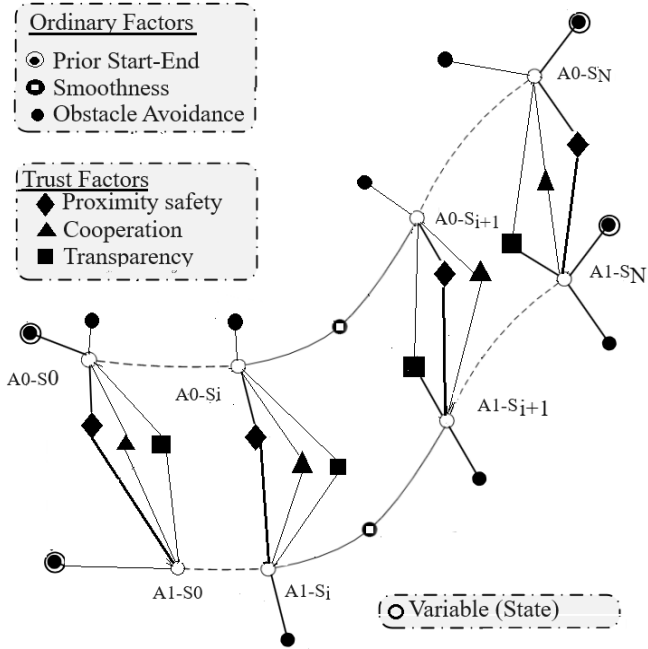

Fig. 2. Factor graph model of proposed method for two typical Ground Robot agents and their interaction using unary and Trust-related factors. As we know in variable-based AI models, we define the problem using a Bayesian formula that includes both prior and likelihood. By accurately defining the prior information and the likelihoods of all events that affect the variables, and ensuring sufficient training, we can identify the variables that satisfy all constraints. The problem can be formulated as a Maximum A Posteriori (MAP) problem as follows:

$$P(\theta|e) \propto P(\theta)L(e|\theta). \quad (1)$$

$P(\theta)$ shows the GP prior and $L(e|\theta)$ is the likelihood of observed events given the variable vector(state) of $\theta$. $L(e|\theta)$ can be computed as the product of all factors including the trust related factors. Identifying the exponential family functions for both the prior and likelihoods will facilitate faster convergence and enhance sparsity.

$$P(\theta|e) \propto \prod_{f=1}^{N_f} factor_{e_f}(\theta), \quad (2)$$

This is a MAP problem that can be illustrated with a factor graph. Considering the exponential family for the events' cost exploits sparsity to solve $P(\theta|e)$ in linear time. To model the interpolation and uncertainty, the linear time-varying stochastic differential equation (LTV-SDE) form of GP needs to be used [20] [21] [2].

*B. GP Prior Factor*

In the context of factor graphs, for linearization and sparsity objectives, we prefer using factors as constraints. These are typically represented by a loss function embedded in an exponential family distribution. Factors related exclusively to θ can be categorized in this section. Dynamics of the GP are centered around the mean. The regular priority function for this purpose can be expressed as follows:

$$P(\theta) \propto \exp\left\{-\frac{1}{2} \| \theta - \mu \|_K^2\right\}, \quad (3)$$

where $\|.\|_K^2$ is the normalized distance with hyperparameter K.

*C. Factor of Unary Events*

In our variable-based model, events that are related to only one variable are considered in this category. For instance, the likelihood function for obstacle avoidance represents the probability of avoiding collisions. All these likelihood functions are defined as distributions within the exponential family. We can categorize different factors, such as regular unary factors, which are only related to one variable. It is important to note that all N variables in the entire trajectory must be considered with respect to that factor. The general formula for a unary factor likelihood is described as follows:

$$L_{e_u}(\theta; e_u) \propto \exp\left\{-\frac{1}{2} \| h_u(\theta) \|_{\Sigma_{e_u}}^2\right\}, \quad (4)$$

where $h_u(\theta)$ is a Hinge loss function for a given current variable (process role) $\theta$, $e_u$ is the corresponding unary event, and $\Sigma_{e_u}$ is the hyperparameter of distribution [22]. The Hinge loss can be written as

$$h_u(\theta) = c_{e_u}(x(\theta, S)), \quad (5)$$

where $x$ is the forward kinematics for shape $S$, $c$ is the Hinge loss function. For example, for the obstacle avoidance factor, the Hinge loss for obstacles can be written as:

$$c_{Obs}(z) = \begin{cases} -d(z) + \epsilon & if\ d(z) \leq \epsilon \\ 0 & if\ d(z) > \epsilon \end{cases}, \quad (6)$$

where $d(z)$ is the signed distance from any kinematic $z$ in $\theta$ to the closest obstacle surface, and $\epsilon$ is a 'safety distance' indicating the boundary of the 'unsafe area' near obstacle surfaces. The signed distance $d(z)$ is computed based on an environment map.

*D. Trust Related Events*

In this project, we focus on three interdependent factors for trust: Proximity Safety, along with Cooperation and Transparency. Additional factors relevant to the application can be incorporated in a similar way to address different problems.

Note that we assume the existence of a peer-to-peer network among agents for message passing, which has been utilized to compute interdependent costs within a standard communication model. Other methods for coding, encrypting, and securing this communication will be considered in future work.

For a given configuration $\theta, e_t$, the likelihood of an inter-dependent event can be written as

$$L_{e_t}(\theta_a; e_t) \propto \prod_{\substack{a'=1 \\ a' \neq a}}^{m} \exp\left\{-\frac{1}{2} \| g_t(\theta_a, \theta_{a'}) \|_{\Sigma_t}^2\right\}, \quad (7)$$

where $g_t(\theta_a, \theta_{a'})$ is a function that defines the cost of two agents behavior $\theta_a$ and $\theta_{a'}$ interfering with each other, $m$ is the number of agents and $\Sigma_t$ is a hyperparameter of the distribution for the inter-dependent factor. This interfering cost function $g_t(\theta_a, \theta_{a'})$ for each interdependent factor can be calculated in a different way. For example, the proximity safety cost for two robots can be calculated based on the minimum distance between the vehicles in relation to a specified threshold, Eqs. (8) and (9).

$$g_t(\theta_a, \theta_{a'}) = c_t(x(\theta_a, S_a), x(\theta_{a'}, S_{a'})), \quad (8)$$

where $x(\theta_a, S_a)$ represents the forward dynamics of agent $a$ in relation to the robot's shape $S_a$. The Hing loss of interaction between two agents to maintain a safe distance can be computed as follows:

$$c_t(z, z') = \begin{cases} -d(z, z') + \epsilon_t & if\ d(z, z') \leq \epsilon_t \\ 0 & if\ d(z, z') > \epsilon_t \end{cases}, \quad (9)$$

Where $d(z, z')$ is the distance between two dynamics, and $\epsilon_t$ is the epsilon distance.

The subsequent set of factors pertains to costs associated with consistency (cooperation) and transparency. These factors are evaluated by comparing the process roles of the truster and the trustee. In our project, particularly in the context of trust, we focus on Behavioral Consistency. This concept refers to the uniformity in agents' behaviors across identical scenarios. To assess consistency, our method involves defining each point using stochastic variables that represent states, such as position and velocity. Acceleration, a crucial metric in this context, is calculated using two adjacent variables. The cooperation cost, for instance, may be determined based on the similarities in acceleration patterns. More intricate models for measuring consistency are planned for future research.

Trust-related transparency in multi-robot path planning plays a vital role, primarily focusing on the reliability and authenticity of the information exchanged between agents. Ensuring transparency is essential for the accuracy and trustworthiness of shared data, especially regarding trajectories and intentions. A key aspect of this is verifying the authenticity of shared trajectories, ensuring they truly represent an agent's intentions and are not fabricated or incorrect. This verification process becomes critical in scenarios where the accuracy of information directly influences the system's safety and efficiency.

Readdressing the factors related to proximity safety, cooperation, and transparency, they are assessed by comparing the process roles of the truster and trustee. Cooperation cost, for example, might be evaluated based on factors like distance and similarities in acceleration. Among collaborating robots, a consistent difference in their accelerations is typically anticipated. Moreover, the transparency cost entails analyzing the discrepancies between the beliefs of agents i and j regarding each other's positions, denoted as θij and θji. The transparency factor is somewhat different from proximity safety and consistency. In proximity safety and consistency, an agent can adjust its behavior to reduce the risk of collision or inconsistency. However, in the case of transparency, the approach is different. Here, we assess the level of transparency by comparing the differences between our beliefs and the information provided by other agents. Our objective in this context is not to gain advantages but to minimize potential costs or risks associated with misinformation. The key aspect of managing transparency is reducing the likelihood of failure by adjusting the thresholds based on the specific agent's behavior and the degree of discrepancy in the information they provide. In simpler scenarios, we can calculate a discrepancy factor and tailor our decision-making processes accordingly, using updated safety thresholds to ensure more reliable and effective responses. The practical implementation of these interconnected factors is detailed in Section IV.

The final optimization problem is expressed in equation (2). To solve this optimization problem, it can be transformed into a linear system, allowing for an iterative solution. Substituting the factors from Eqs. (3), (4), and (7) into Eq. (2) results in the following equation:

$$P(\theta|e) = P(\theta) \prod_{u=0}^{N_u} L_{e_u}(\theta; e_u) \prod_{t=0}^{N_t} L_{e_t}(\theta_a; e_t) \quad (8)$$

Where $N_u$ and $N_t$ represent the number of unary factors and trust related factors, respectively. This scenario resembles a well-known MAP problem that can be solved using iterative algorithms such as Gauss-Newton or Levenberg-Marquardt, continuing until convergence.

$$\theta^* = \underset{\theta}{argmin} \Big\{ \|\theta - \mu\|_K^2 + \|h_{obs}(\theta)\|_{\Sigma_{obs}}^2 + \frac{1}{2}\sum_{a'=1}^{m} \| \quad (9)$$
$$g_{t_1}(\theta_a, \theta_{a'})\|_{\Sigma_{t_1}}^2 + \frac{1}{2}\sum_{a'=1}^{m} \| g_{t_2}(\theta_a, \theta_{a'})\|_{\Sigma_{t_2}}^2 +$$
$$\frac{1}{2}\sum_{a'=1}^{m} \| g_{t_3}(\theta_a, \theta_{a'})\|_{\Sigma_{t_3}}^2 \Big\},$$

where $h_{obs}, g_{t_1}, g_{t_2}$ and $g_{t_3}$ represent costs related to obstacles, proximity safety, consistency, and transparency, respectively.

The optimization problem in Eq. (9) becomes polynomial, whose derivative is easy to compute. The corresponding linear equation based on the Gaussian-Newton method is as follows:

$$\theta^* = \underset{\theta}{argmin} \| A\theta - b \|^2, \quad (10)$$

where $A \in \mathbb{R}^{n_f \times n_s}$ is the measurement Jacobian consisting of $n_f$ measurement rows and $b$ is an $n_s$-dimensional vector computable similar to iSAM2 [23].

## IV. UNSIGNALIZED INTERSECTION PROBLEM

An unsignalized intersection refers to any junction of two or more public roads where traffic signal control is absent, allowing vehicles to navigate based on established right-of-way rules. These intersections can be classified using a variety of criteria. For instance, they can be categorized by the type of traffic control mechanism employed, such as stop signs or yield signs. Additionally, classifications can be based on the surrounding area's characteristics, encompassing designations like residential, commercial, industrial, or rural. Furthermore, intersections can be differentiated by the number of approaches they possess, including options like three-way or four-way configurations.

In the context of this experiment, our focus lies on a four-way intersection in an urban area where vehicles in the area collaborate through inter-vehicle communications. We do not take right of way into account; instead, we focus on three standard factors in agent behavior: prior factors, smoothness, obstacle avoidance, and three interdependent trust-related factors, namely proximity safety, consistency, and transparency. The optimization process takes into account well-defined destinations and aims to determine a trustworthy path. Vehicles capable of establishing communication form collaborative groups, enabling the sharing of their respective process role in the overall agents. To simplify matters, we assume instantaneous access to a precise snapshot of the environment's state; otherwise, agents would need to estimate and reconcile each other's beliefs. In future works, we aim to explore alternative algorithms that can decrease the traffic network between agents without sharing the entire trajectory. Additionally, instead of assuming the availability of a full map, we will consider using a map with partial observability.

This trustworthy behavior is vividly illustrated through factor graph that guide the vehicles' movements. These trajectories are represented by a vector value function, containing limited number of positions and corresponding velocities. Individual vehicles are empowered to make decisions and ascertain their optimal behavior by factoring in GP prior, unary, and trust related inter-agent factor.

Among the unary factors discussed in this paper are smoothness and obstacle avoidance, while the trust-dependent inter-agent factors include proximity safety, cooperation, and transparency. To simulate real-world scenarios, we developed a system featuring four cars arriving at an intersection, each with predetermined destinations, as illustrated in Fig. 3. In our simulated environment, which spans a 5 by 5 meter area, we have deployed four robots: Alpha, Beta, Gamma, and Delta. Each robot is uniquely identified by its own color and has a circular shape with a diameter of 20 cm. For safety and operational efficiency, we ensure a minimum clearance of 10 cm between the robots and any obstacles present in the environment (Black areas in Fig. 3).

*A. Proximity Safety Test*

In our initial scenario, as illustrated in Fig. 3, we conducted a simulation involving four labeled robots, each programmed with a specific destination. This simulation was executed in two distinct phases: the first phase did not incorporate trust factors, and the second phase operated completely devoid of any trust considerations. The setup for both phases was identical to the environment depicted in Fig. 3. In this test we focused on evaluating the proximity safety factor in each phase.

In the phase without trust factors, our findings (presented in Fig. 4(B)) revealed a proximity safety breach between the blue and red robots, highlighted by a circle in the figure. The detailed interference matrix pertaining to this phase is depicted in Fig. 5. Conversely, the second phase of the simulation, as depicted in Fig. 4(A), showed a marked improvement where the average safety factor consistently exceeded the minimum required distance between the agents.

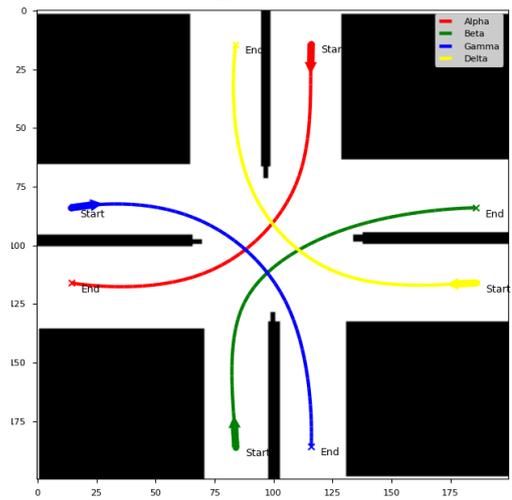

**Fig. 3.** Unsignalized intersection with four robots with specified sources and destinations

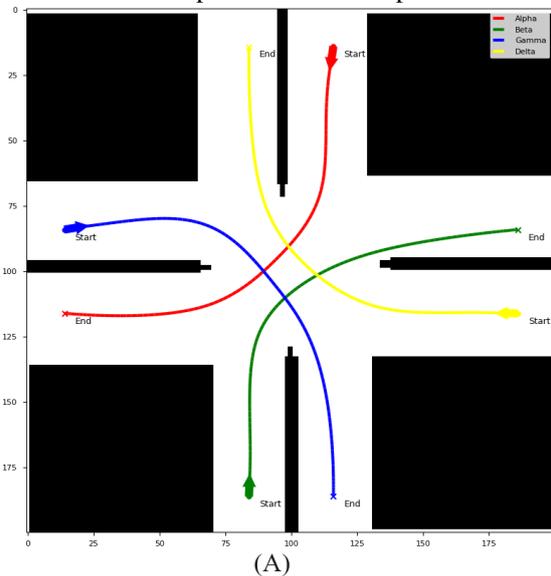
(A)

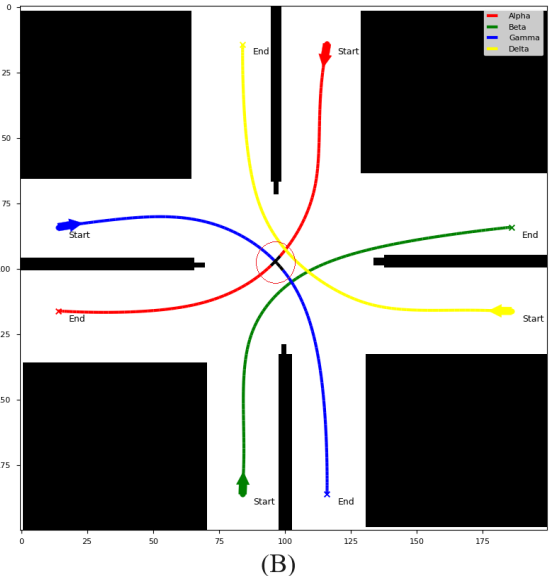
(B)

**Fig. 4.** In subsection (B), the phase that does not include trust factors, a proximity safety breach between the blue and red robots was revealed, which is highlighted by a circle in the figure.

Each trajectory is broken down into segments (each segment is a line between two consecutive points). For each segment, we check if it's too close to any segment of another trajectory. If so, we change its color to black . The proximity threshold defines what 'too close' means. We should set this according to our specific requirements.

The subfigure Fig. 5(B) corresponds to the scenario where trust factors are excluded from our optimization problem. In this depiction, the red boxes highlight instances where the minimum distance between elements falls below the established threshold of 10 cm.

## B. Consistency check

The concept of consistency at unsignalized intersections can be interpreted in various ways. In our project, particularly in the context of trust, we focus on Behavioral Consistency. This pertains to the uniformity of drivers' behaviors in identical intersection scenarios. We have implemented context-based behavior for agents using a GP factor graph, which represents the trajectory. To analyze the consistency features in corresponding trajectories, we employ a method where each point is defined by stochastic variables akin to state, including position and velocity. Acceleration is a crucial metric for assessing consistency, and we calculate it using two adjacent variables. In this study, we analyze the consistency of four robots approaching an intersection by examining the velocity change diagrams to gauge their behavioral similarity, Fig. 6. This analysis is conducted twice — once with consistency trust factors and once without — to compare the results and better understand the impact of trust on consistency. The results indicate approximately a 15% inconsistency in acceleration when the consistency factor was not considered, and only the proximity safety factor was taken into account.

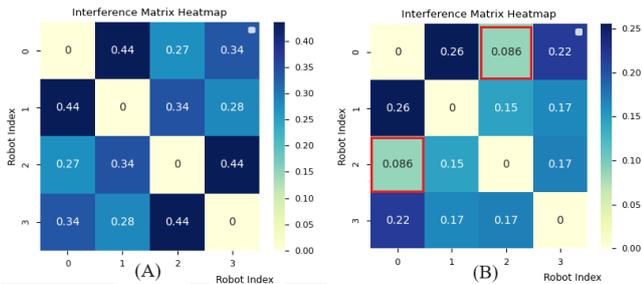

**Fig. 5.** The minimum distance between each pair of trajectories serves as an indicator of interference per meter. In part (A), the distances meet the established requirement of maintaining a 0.1 m threshold. However, in part (B), this threshold is not met, indicating a failure to maintain the required separation.

## C. Transparency check

Trust-related transparency in multi-robot path planning is critically important, primarily focusing on the reliability and authenticity of information exchanged between agents. Such transparency is essential to ensure that the data shared, especially concerning trajectories and intentions, is accurate and trustworthy. A key aspect is verifying the authenticity of shared trajectories. This involves mechanisms to confirm that the trajectories and plans shared by an agent genuinely reflect its intentions and are neither fabricated nor incorrect. This verification process is crucial in scenarios where the accuracy of shared information is directly tied to the system's safety and efficiency.

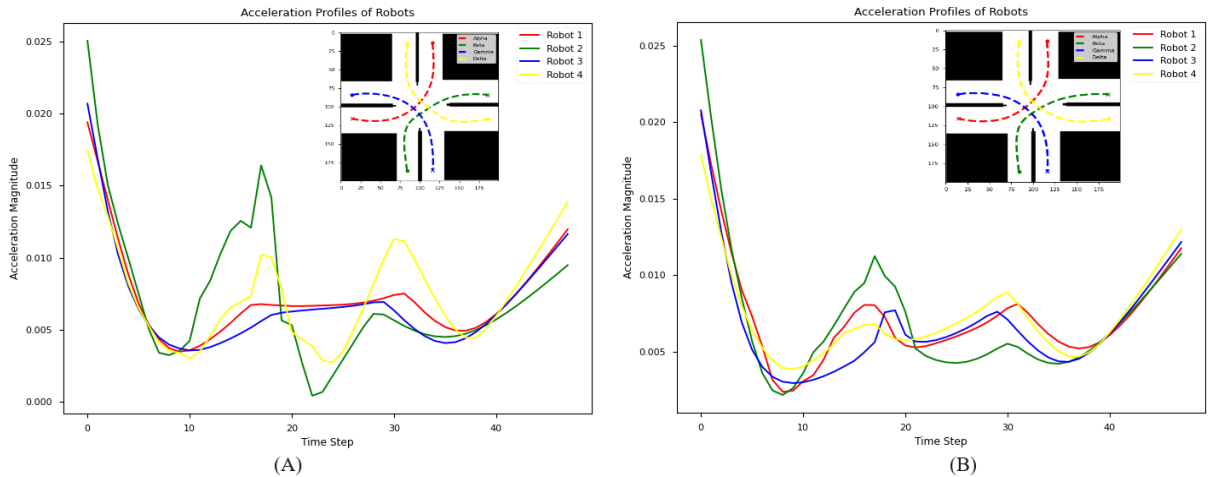

**Fig. 6**. Comparing the consistency of acceleration for the scenario depicted in Fig. 4, with and without considering consistency factors, sub figures B and A respectively, has been illustrated. The results show about 15% inconsistency when the factor is not employed in subfigure A.

In our method, agents share their trajectory data, and decisions are based on this information. However, concerns arise regarding the potential falsification of this data, even with assured authentication. For example, a compromised agent might produce inaccurate information. To counter this, we propose observing the environment, under the assumption of full environmental awareness, to verify the positions of agents in relation to their shared trajectories. Furthermore, as part of our future work, we aim to incorporate partial observability, mapping different partial observations to form a comprehensive belief system. By comparing these beliefs with the received information, we can measure any discrepancies. As previously discussed in Section III.D, to elicit an appropriate response based on the discrepancy factor and to refine our decision-making processes, we have updated safety thresholds, such as the epsilon distance, to ensure more reliable and effective responses.

In practice, we compute the level of transparency for each agent and incorporate this into other interrelated factors, such as proximity safety, by extending the thresholds. Figure 7 demonstrates the effect of transparency checks on one of the agents that does not have transparent data (Beta-Green robot). In fact, we simulated the behavior of one agent to exhibit 40% non-transparency. This led to a noticeable change in the behavior of other agents; they attempted to maintain a greater distance from this particular agent. In Fig. 8, it is evident that the minimum distance maintained when using transparency checks is almost two times greater compared to the 'Beta' robot, which is indexed as 1.

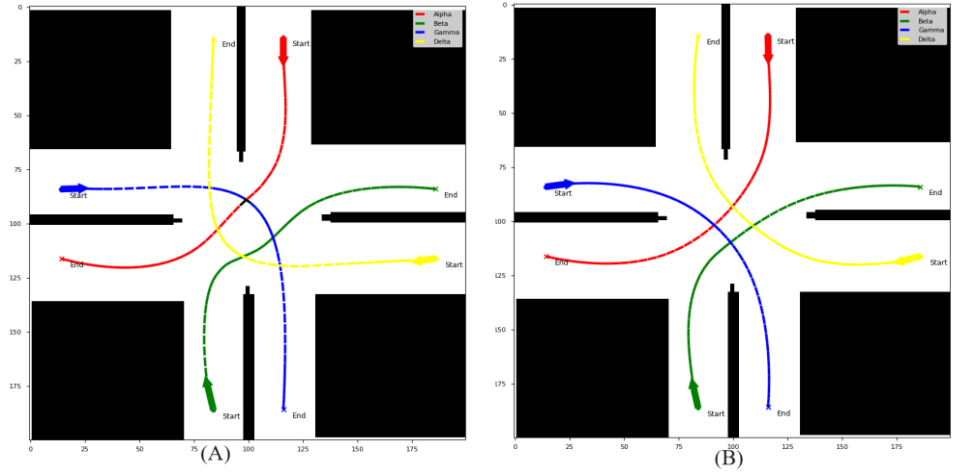

**Fig. 7.** In subsection (B), where transparency is not an issue, as opposed to subsection (A) where the green agent exhibits a 40% failure in transparency, it is noticeable that all agents try to avoid the green agent.

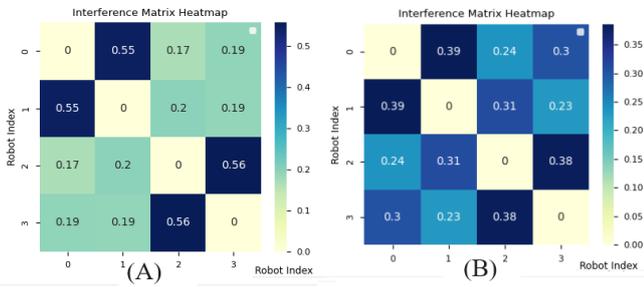

**Fig. 8.** In part (A) of the illustration, the distances between agents in relation to the green agent, indexed by 1, are almost twice as large compared to those in Figure B, where the transparency factor is not considered.

## V. REAL WORLD IMPLEMENTATION

To experimentally validate the effectiveness of the proposed method, four autonomous vehicles were utilized in an indoor, unsignalized intersection environment. The lab cars shared their process roles via Wi-Fi. An Optitrack system, consisting of cameras, was used to capture the motion of the cars and feed it back to the workstation for enhanced observability. Figure 5 illustrates this communication flowchart. The test videos are available at https://youtu.be/zcxs5mAdpO8.

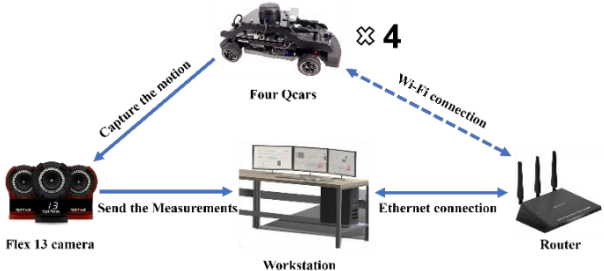

**Fig. 9**. Communication setting in lab

In the tests, we configured all vehicles to make a left turn to assess our algorithm's performance in collision avoidance and trajectory smoothness. The initial position coordinates for car 1 to car 4 were set at [0.4, -2.15], [-0.4, 2.15], [-2.15, -0.4], and [2.15, 0.4], respectively. The corresponding destination coordinates were [-2.15, 0.4], [2.15, -0.4], [0.4, 2.15], and [-0.4, -2.15].

During this experiment, the transparency factor was evaluated. Initially, all vehicles shared 100% authenticated information. However, in scenarios (A) and (B), we introduced low transparency by programming one of the robots, referred to as 'red,' to exhibit 40% misinformation. The resulting trajectories are shown in Figure 10, which demonstrates the adjustments made with increasing epsilon distance thresholds. The outcomes are evident in the minimum distance matrices. In case B, the robots attempt to adjust their distances to compensate for the dishonesty.

Minimum Distance Matrix for Scenario (A).
MinDistance(A):
 [[0, 0.28, 0.20, 0.22],
  [0.28, 0, 0.25, 0.42],
  [0.20, 0.25, 0, 0.48],
  [0.22, 0.42, 0.48, 0]]

Minimum Distance Matrix for Scenario (B):
MinDistance(B):
 [[0, 0.62, 0.31, 0.39],
  [0.62, 0, 0.18, 0.47],
  [0.31, 0.18, 0, 0.33],
  [0.39, 0.47, 0.33, 0]]

As evident from the matrices, the minimum distance for Car 1 in Scenario (B) is significantly higher compared to the previous case.

## VI. CONCLUSIONS

This paper presents the successful implementation of a comprehensive trust management system tailored for multi-robot systems. It integrates trust evaluation and establishment components, as well as consensus, within a single optimization problem. In a decentralized environment, trustworthy decision-making for each agent requires observing its individual behavior while also considering information from other agents. To enable this holistic assessment, we introduce a graphical model using factor graphs and an innovative Bayesian inference methodology. This approach models trustworthy behavior and finds the optimal process roles with consent. It incorporates essential factors: unary factors necessary for path planning, such as smoothness and obstacle avoidance, and trust-related inter-agent factors like proximity safety, cooperation, and transparency. Collectively, these factors contribute to trustworthy behavior. Our algorithms enforce consensus among multiple agents as they share and adapt based on the received data.

For future work, this novel approach holds potential for other applications, such as surveillance systems and environments where robots and humans interact with each other.

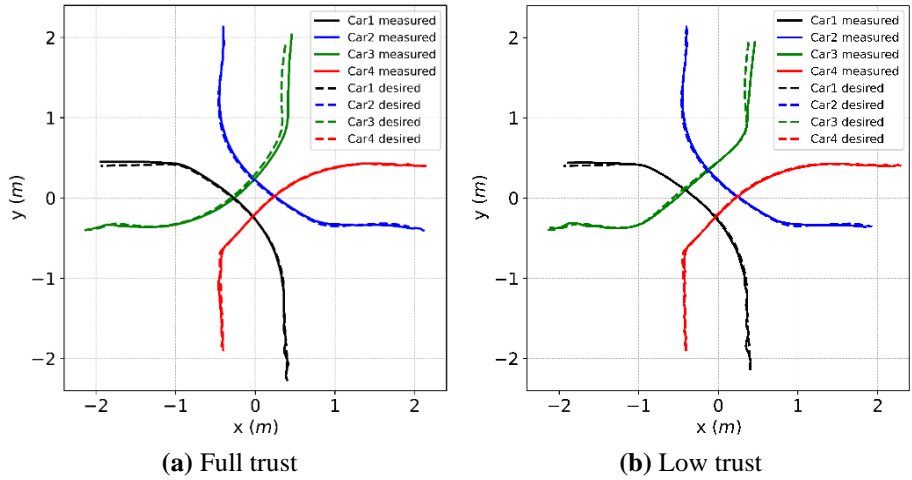

**(a)** Full trust      **(b)** Low trust

**Fig. 10.** Unsignalized intersection final paths for two inter trust matrices (A) and (B).

ACKNOWLEDGMENT

This work has been partially supported by the Natural Sciences and Engineering Research Council of Canada (NSERC) through its Discovery Grant (RGPIN 04818-2018), and by the Social Sciences and Humanities Research Council of Canada (SSHRC) through its Insight Grant (IG 435-2023-1056). Additional support comes from the Innovation for Defence Excellence and Security (IDEaS) program of the Canadian Department of National Defence (DND) CFPMN2-051-York University. The opinions and conclusions presented in this work are solely those of the author(s) and should not be construed as reflecting the views, positions, or policies of IDEaS, DND, or the Government of Canada.

REFERENCES

[1] C. E. Rasmussen and C. K. I. Williams, Gaussian Processes for Machine Learning, MIT Press, 2006.

[2] M. Mustafa, Y. Xinyan and B. Byron, "Gaussian Process Motion Planning," in *Proc. IEEE Conference on Robotics and Automation (ICRA-2016)*, 2016.

[3] J.-S. Ha, H.-J. Chae and H.-L. Choi, "Approximate Inference-Based Motion Planning by Learning and Exploiting Low-Dimensional Latent Variable Models," *IEEE Robotics and Automation letters*, vol. 3, no. 4, pp. 3892-3899, October 2018.

[4] R. Ismail and A. Jøsang, "The Beta Reputation System," in *Bled eConference*, 2002.

[5] S. Buchegger and J.-Y. L. Boudec, "A Robust Reputation System for Mobile Ad-hoc Networks," in *EPFL IC Technical Report IC*, 2003.

[6] W. T. L. Teacy, J. Patel, N. R. Jennings and M. Luck, vol. 12, no. 2, p. 183–198, 2006.

[7] J. Zhang and R. Cohen, "Evaluating the trustworthiness of advice about seller agents in e-marketplaces: A personalized approach," *Electronic Commerce Research and Applications,* vol. 7, no. 3, pp. 330-340, September 2008.

[8] N. Griffiths, "Task Delegation using Experience-Based Multi-Dimensional Trust," in *4th Int. Joint Conf. Auto. Agents Multiagent Syst*, 2005.

[9] C. Castelfranchi, R. Falcone and G. Pezzulo, "Trust in Information Sources as a Source for Trust: A Fuzzy Approach," in *2nd Int. Joint Conf. Auto.*, 2003.

[10] U. Kuter and J. Golbeck, "SUNNY: A New Algorithm for Trust Inference in Social Networks Using Probabilistic Confidence Models," in *22nd Nat. Conf. Artif. Intell.*, 2007.

[11] M. E. G. Moe, M. Tavakolifard and S. J. Knapskog, "Learning Trust in Dynamic Multiagent Environments using HMMs," NordSec, 2008.

[12] E. ElSalamouny, V. Sassone and M. Nielsen, *Formal Aspects in Security and Trust*, vol. 0, no. 9, p. 21–35, 2009.

[13] W. T. L. Teacy, G. Chalkiadakis, A. Rogers and N. R. Jennings, "Sequential decision making with untrustworthy service providers," in *7th Int. Joint Conf. Auto. Agents Multiagent Syst*, 2008.

[14] H. Zhu, E-CARGO and Role-Based Collaboration: Modeling and Solving Problems in the Complex World, NJ, USA: Wiley-IEEE Press, Dec. 2021.

[15] S. Russell and P. Norvig, Artificial Intelligence: A Modern Approach, Upper Saddle River, N.J.: Prentice Hall, 2022.

[16] Y. Ni, D. Jones and Z. Wang, "Consensus Variational and Monte Carlo Algorithms for Bayesian Nonparametric Clustering," in *IEEE International Conference on Big Data (Big Data)*, Atlanta, GA, USA, 2020.

[17] E. Bellini, Y. Iraqi and E. Damiani, "Blockchain-Based Distributed Trust and Reputation Management Systems: A Survey," *IEEE Access,* vol. 8, pp. 21127-21151, January 27, 2020.

[18] A. Baliga, "Understanding blockchain consensus models," *Persistent,* no. 4, p. 1–14, 2017.

[19] S. Bandyopadhyay and S.-J. Chung, "Distributed Estimation using Bayesian Consensus Filtering," *arXiv*, p. 10.48550/ARXIV.1403.3117, 2014; https://arxiv.org/abs/1403.3117.

[20] B. Akbari, Z. Wang, H. Zhu, L. Wan, R. Adderson and Y. -J. Pan, "Role Engine Implementation for a Continuous and Collaborative Multirobot System," *IEEE Transactions on Systems, Man, and Cybernetics: Systems,* 2023.

[21] B. Akbari and H. Zhu, "Fault-Resilience Role Engine for an Autonomous Cooperative Multi-Robot System using E-CARGO," in *IEEE Int'l Conf. on Systems, Man, and Cybernetics*, Prague, Czech, Oct. 2022.

[22] M. Zucker, N. Ratliff, A. Dragan, M. Pivtoraiko, M. Klingensmith, C. Dellin, J. Bagnell and S. Srinivasa, "CHOMP:Covariant Hamiltonian optimization for motion planning," *The International Journal of Robotics Research,* vol. 32, no. 1164–1193, pp. 9-10, Sep. 2013.

[23] M. Kaess, H. Johannsson, R. Roberts, V. Ila, J. Leonard and F. Dellaert, "iSAM2: Incremental Smoothing and Mapping Using the Bayes Tree," *The International Journal of Robotics research.,* vol. 31, no. 2, pp. 216-235, May 2012.

[24] S. Anderson, T. D. Barfoot, C. H. Tong and S. Särkkä, "Batch Continuous-Time Trajectory Estimation as Exactly Sparse Gaussian Process Regression," *Autonomous Robots,* vol. 39, no. 3, pp. 221-238, July 2015.